\documentclass{article}

\usepackage{arxiv}

\usepackage[utf8]{inputenc} 
\usepackage[T1]{fontenc}    
\usepackage{hyperref}       
\usepackage{url}            
\usepackage{booktabs}       
\usepackage{amsfonts}       
\usepackage{nicefrac}       
\usepackage{microtype}      

\usepackage{graphics}
\usepackage{epsfig}
\usepackage{caption}
\usepackage{subcaption} 
\usepackage{cite}
\usepackage{amsmath}
\usepackage{amssymb}

\title{Human-centered collaborative robots with deep reinforcement learning}

\author{
  Ali Ghadirzadeh$^*$ \\
  \texttt{algh@kth.se}\\
  \And
  Xi Chen$^*$\\
  \texttt{xi8@kth.se}\\
  \And
  Wenjie Yin\\
  \texttt{yinw@kth.se}\\
  \And
  Zhengrong Yi\\
  \texttt{zhyi@kth.se}\\
  \AND
  M{\aa}rten Bj{\"o}rkman\\
  \texttt{celle@kth.se}\\
  \And
  Danica Kragic\\
  \texttt{dani@kth.se}\\
  \AND
  Division of Robotics, Perception and Learning (RPL)\\
  KTH Royal Institute of Technology \\
  \\$^*$The authors contributed equally \\
}

\begin{document}
\maketitle

\begin{abstract}
We present a reinforcement learning based framework for human-centered collaborative systems. The framework is proactive and balances the benefits of timely actions with the risk of taking improper actions by minimizing the total time spent to complete the task. 
The framework is learned end-to-end in an unsupervised fashion addressing the perception uncertainties and decision making in an integrated manner. The framework is shown to provide more fluent coordination between human and robot partners on an example task of packaging compared to alternatives for which perception and decision-making systems are learned independently, using supervised learning. The foremost benefit of the proposed approach is that it allows for fast adaptation to new human partners and tasks since tedious annotation of motion data is avoided and the learning is performed on-line.
\end{abstract}

\keywords{Human-robot Collaboration \and Deep Reinforcement Learning}

\section{Introduction}
\label{sec:introduction}

Human-centered collaborative systems require proactive robot behavior with precise timing, 
which in turn mandates awareness of human actions, state of the environment and the task being executed, \cite{ajoudani2018progress, amershi2019guidelines, zhang2020recurrent, cheng2020towards}. 
Proactive robot behavior is  achieved by (1) recognizing the current state of the human collaborator and the environment based on real-time observations, (2) human action prediction given the observations and the model of the task, and (3) generating robot actions in line with the prediction.

Human action recognition may however be highly uncertain if the human collaborator is not executing a strictly defined task plan. 
This is true regardless of whether perception is based on motion-capture devices or image based pose estimation. For a robot to act in a proactive manner, while at the same time avoiding actions when the risk of making a mistake is too high, it is essential for the action-decision system to take this uncertainty into consideration.

We therefore propose to train the perception system and the robot policy in an end-to-end fashion using reinforcement learning (RL). This is different from earlier studies in which human action recognition and prediction are typically decoupled from robot action policy training \cite{zhang2020recurrent,cheng2020towards, kwon2014planning, hawkins2014anticipating, huang2016anticipatory}. Our main objective is to improve the 
fluency in coordination between the human and robot partners by allowing the policy to explicitly weigh the benefits of timely actions to the risk of making a mistake when uncertainties are too high.

We introduce a deep RL framework to obtain proactive action-selection policies 
implemented as a state-action value function (in short, value function) that receives full-body motion data from a motion capture suit and outputs the value of performing each action given the current state of the task.
As the main contribution of this work, we demonstrate that the coordination between the human-robot pair improves when the perception (recognizing and predicting human behavior) and the policy (robot action-decisions) are trained jointly to optimize the time-efficiency of the task execution. 
The benefits of our approach compared to the earlier work are 
(1) improving robot action decision making by an efficient handling of the uncertainties in human action recognition considering a collaborative setup, (2) enabling the robot to distinguish whether it is optimal or not to take a proactive action, and (3) eliminating the need for tedious manual labeling of human activities by learning directly from raw sensory data. 

\begin{figure}[!t]
\centering
\includegraphics[width=0.5\linewidth]{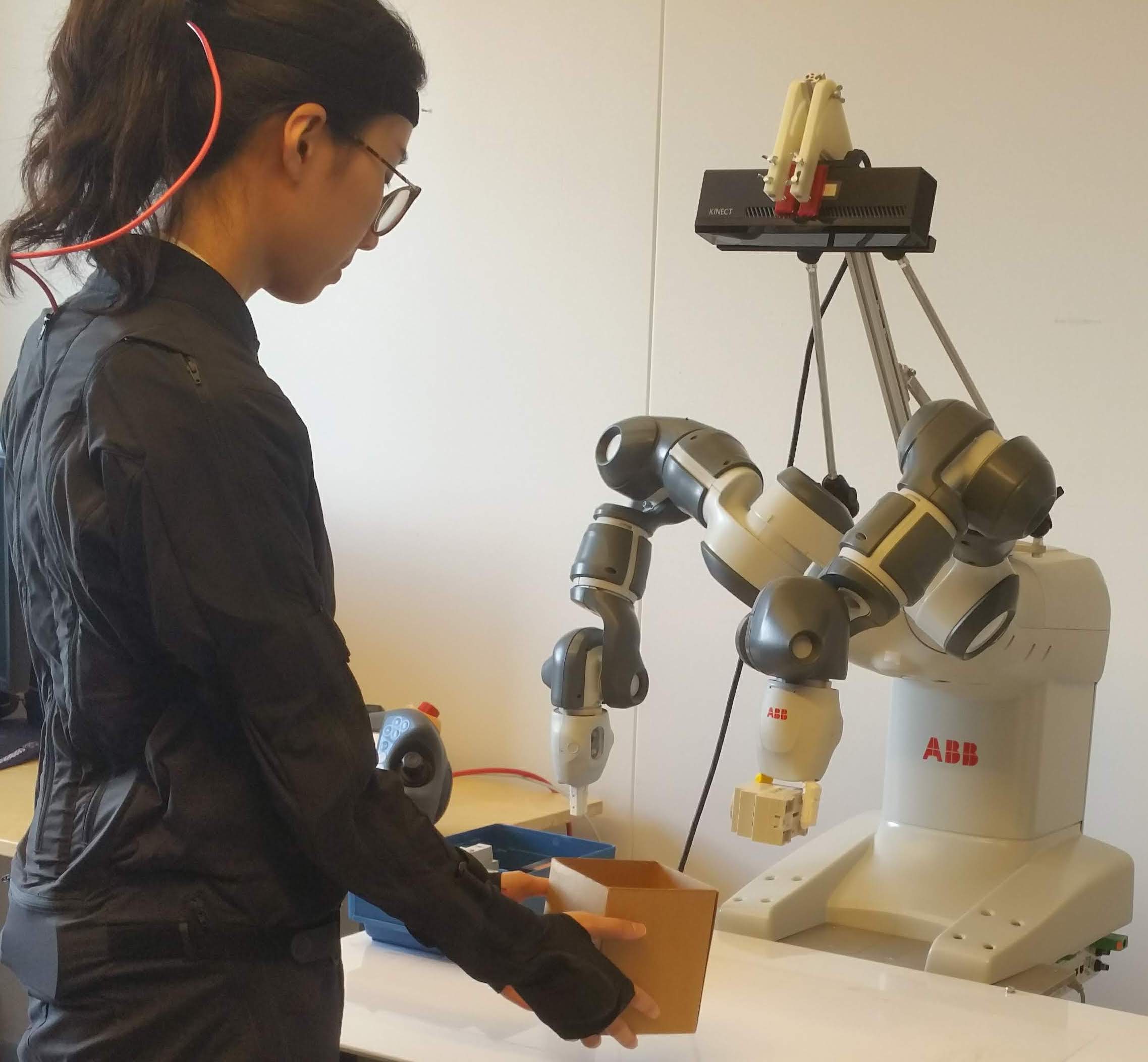}
\caption{The collaborative packaging task setup. A person wearing a motion capture suit brings a box to the robot. By analyzing the motion pattern of the person while handling the box, the robot can proactively pick up the appropriate item before it observes the type of the box.}
\label{fig:setup}
\end{figure}

We exploit graph convolutional networks (GCNs) \cite{yan2018spatial} and recurrent Q-learning \cite{hausknecht2015deep} to process the sequential motion data. To enhance data efficiency for the learning framework, we train an auxiliary unsupervised motion reconstruction network that learns a representation used by the value function. 
We also propose the use of behavior trees \cite{colledanchise2016behavior} as a means to structure the prior knowledge of the task.
We evaluate the proposed learning framework on a collaborative packaging task shown in Fig.~\ref{fig:setup}, and show how the fluency of the task coordination in human-robot collaboration scenarios is achieved. 

\subsection{Example scenario}
\label{sec:motivation}
\begin{figure}[!ht]
\centering
\includegraphics[width=0.8\linewidth]{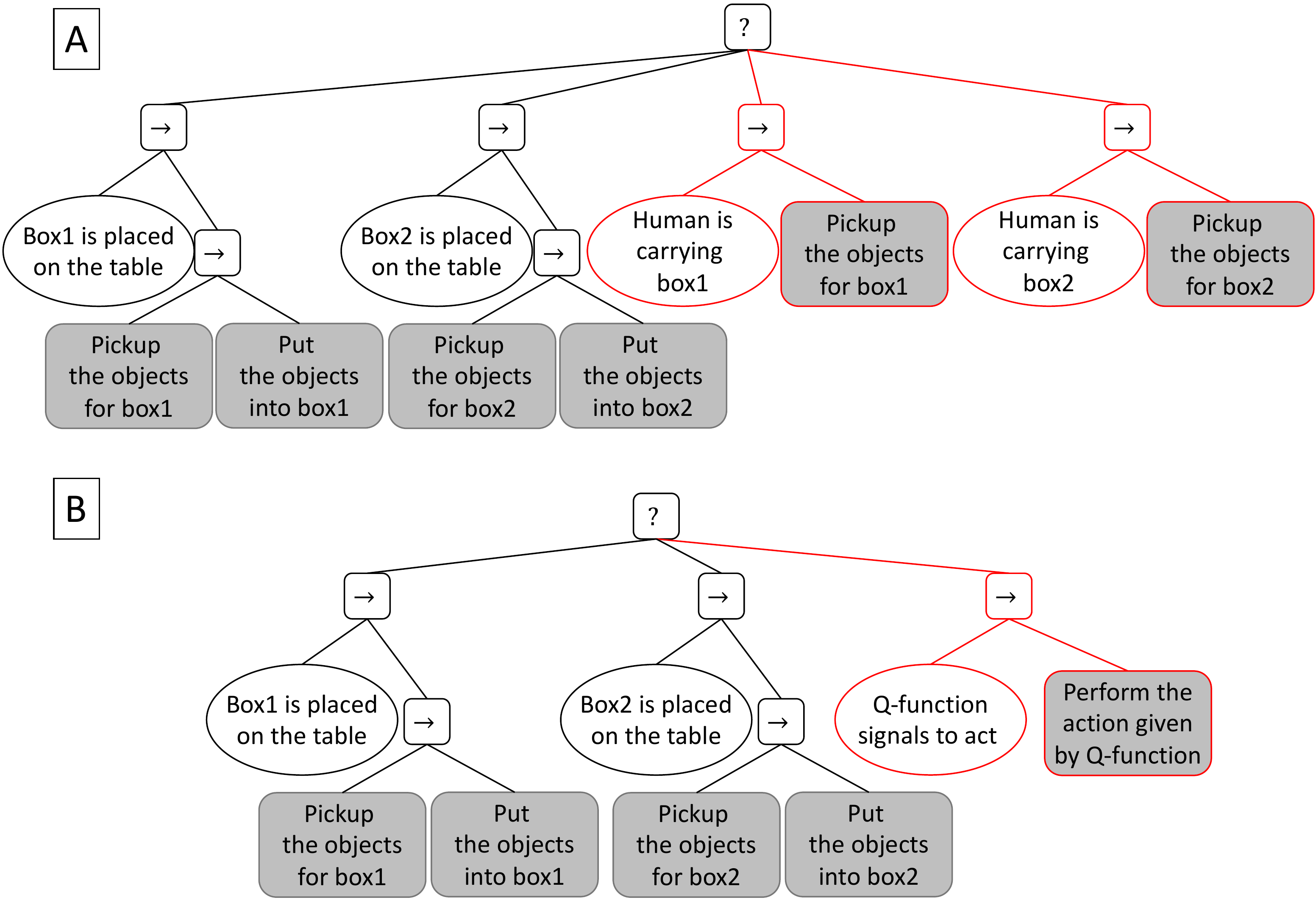}
\caption{The behavior tree of the example scenario.}
\label{fig:bt1}
\end{figure}

We further motivate our work by introducing an example scenario of collaborative human-robot packaging and Fig.~\ref{fig:bt1}A illustrates a typical behavior tree for such a task. In this example, there are two types of boxes that the human partner can arbitrarily choose. A barcode scanner first identifies the type of the box when the box is placed on the table in front of the robot.  This information is then sent to the robot to place the appropriate items into the box.

The nodes with black boundaries in the behavior tree illustrate a minimal execution plan for the robot to complete this task in collaboration with the human partner. However, this plan adds unnecessary delays to the task execution. The robot starts picking up the objects first after it has received the box type information, while it could have gathered this information well before the box is placed on the table. This problem could be addressed by adding the red nodes to the behavior tree, as illustrated in the figure. However, conditions like "person is carrying box1" cannot be evaluated with certainty when the barcode has to be read from a distance, which is also true for human action recognition from raw motion data.

Therefore, we propose to instead add a value function node, shown in Fig.~\ref{fig:bt1}B in red, a node that continuously integrates raw motion data by its recurrent structure, and predicts the right time to act by following a trial-and-error approach to optimize the parameters of the value function. 
Actions can then be executed as soon as the potential benefits of timely assistance outweigh the risk of incorrectly acting on incomplete observations, which in the end provides more fluent coordination with the human partner.

This paper is organized as follows: In the next section, we review the related work. Sec.~\ref{sec:method} describes the details of the proposed learning framework. We provide our experimental results in Sec.~\ref{sec:experiments}, and finally we conclude the paper and introduce our future work in Sec.~\ref{sec:conclusion}. 
\section{Related work}
\label{sec:related_work}

Research on human-robot collaborative (HRC) systems is motivated by its potential for applications in manufacturing, healthcare, and social robotics. Increasing attention has been  put on development of intuitive and seamless HRC systems where human intention is recognized to allow for adapting robot behavior in real-time. Thus, human body movement prediction\cite{butepage2020imitating}, gaze and gesture recognition\cite{sibirtseva2019exploring} have been put forward as more intuitive means for collaboration in comparison to other cues \cite{ajoudani2018progress}, e.g., auditory, force/pressure \cite{ghadirzadeh2016sensorimotor}, bio-signals, etc. We review relevant work based on the three problems considered in this work in the scope of human-robot collaborative systems: human action recognition, dealing with and modeling the uncertainty, and robot action planning and execution. 

Human action recognition has been investigated broadly in the computer vision community for a wide range of applications. In the domain of HRC, human action recognition has been realized using a variety of methods, such as Gaussian mixture models (GMM) \cite{mainprice2013human}, and interaction Probabilistic Movement Primitives (ProMPs) \cite{maeda2017probabilistic}. More recent approaches typically rely on some form of deep learning, using e.g. deep CNNs such as in \cite{wang2018deep}. Graph convolutional networks (GCNs) have been proposed as a means to deal with data in graph structures and have become powerful tools for skeleton based action recognition, especially when extended to spatial-temporal graph convolutional  networks (ST-GCN) \cite{yan2018spatial}, networks that can learn both spatial and temporal patterns in movement data, something that will be exploited in this paper. 

Our focus in this work is on handling the uncertainty in human action recognition and prediction. In HRC systems, one of the common ways to handle uncertainty is to accompany the prediction with a confidence value. A measure of prediction confidence was proposed in \cite{thobbi2011using} for obtaining a weighed combination of reactive and proactive actions. In \cite{huang2016anticipatory}, both prediction and corresponding confidence produced by the classifier are considered in the motion planner with fixed thresholds to proactively plan and execute robot motion. There are also probabilistic approaches to handle uncertainty. In \cite{kwon2014planning}, the probabilistic decision problem was solved by augmenting a Bayesian network with temporal decision nodes and utility functions, enabling simultaneous determination of the nature and time of a proactive action. Bayesian networks were employed to model human motion sequences in \cite{hawkins2013probabilistic,hawkins2014anticipating}, which also presented a cost-based planning algorithm to optimize robot motion timings. Apart from these efforts, Monte-Carlo dropout was incorporated into an RNN-based model \cite{zhang2020recurrent} for uncertainty estimation when predicting human motion trajectories. 
Our approach to handle the uncertainties in HRC is radically different from most prior work in that we train a RL policy to make efficient action-decisions without explicit modeling of the uncertainty. 

For the third problem considered in this work, robot action planning, research generally falls in two categories: (1) low-level motion planning of the robot, and (2) high-level task planning. Work on motion planning, in the context of HRC, addresses the computation of collision-free robot trajectories given human motion data, mostly for the safety reasons \cite{perez2015fast,park2017intention}, and to synchronize the robot motion with the human, e.g., in a hand-over task \cite{zhang2020recurrent}. Our work focuses on the high-level task planning that aims at efficient task representation and execution. 
High-level action planning is typically paired with the ability of the robot to recognize the intention of the human partner \cite{zhang2020recurrent, cheng2020towards, park2017intention}.
However, as demonstrated in this paper, time-efficiency in HRC can be improved when human action prediction and robot action planning are trained jointly in an end-to-end fashion. 
End-to-end reinforcement learning, and more specifically Q-learning, is used in some works to obtain high-level robot action-selection policies to interact with humans in social scenarios \cite{qureshi2016robot} and collaborative tasks  \cite{koppula2016anticipatory} without addressing the time-efficiency aspects for HRC. 
Here, we introduce a deep recurrent Q-learning algorithm that processes long-term human motion data to improve action-decisions to be made more proactively, thus improving the time-efficiency of the task execution. 
\section{Learning framework}
\label{sec:method}
In this section, we introduce our learning framework to obtain an action-selection policy $a_i \leftarrow \pi(  o_0, ..., o_{N \times i} , bt_i ; \theta_\pi)$ that continuously processes a sequence of human motion data $[o_0, ..., o_{N \times i}]$ and the current state of the behavior tree $ bt_i$, and sequentially makes proactive action-decisions $a_i$. $N$ is a positive integer that determines the intervals at which action decisions are made, and $\theta_\pi$ denotes the parameters of the policy network. 

We deal with a partially observable Markov decision process (POMDP) in which the state of the task cannot be directly observed by the learning agent.
The observations can be any temporal measure of human body from different sensing modalities, e.g., body motion data or gaze information that comprise the information required to estimate the state of the person in the collaborative task. In this paper, we only consider body motion data captured by a motion capture suit. 
Actions are sampled from an action set $a_i \in \mathcal{A}$ consisting of , e.g., \textit{"pick up an object"} or \textit{"wait to collect more data"}.

Here, an episodic task is terminated after $T$ time-steps and a reward $r_T$ is only provided at the end of the episode. 
The signal to terminate the episode originates from the behavior tree when any other node asks the robot to perform a task. 
In case the robot has performed an  action that is in line with the given task, a positive reward is provided to the RL agent that is equal to the amount of time that is saved by the proactive behavior of the robot.
On the other hand, if the action performed is unsuitable for the task at the current state,
then it is punished by the amount of delays added to the task execution to revoke the improper action. 

\subsection{Network architecture}
\begin{figure}[h]
\centering
\includegraphics[width=0.7\linewidth]{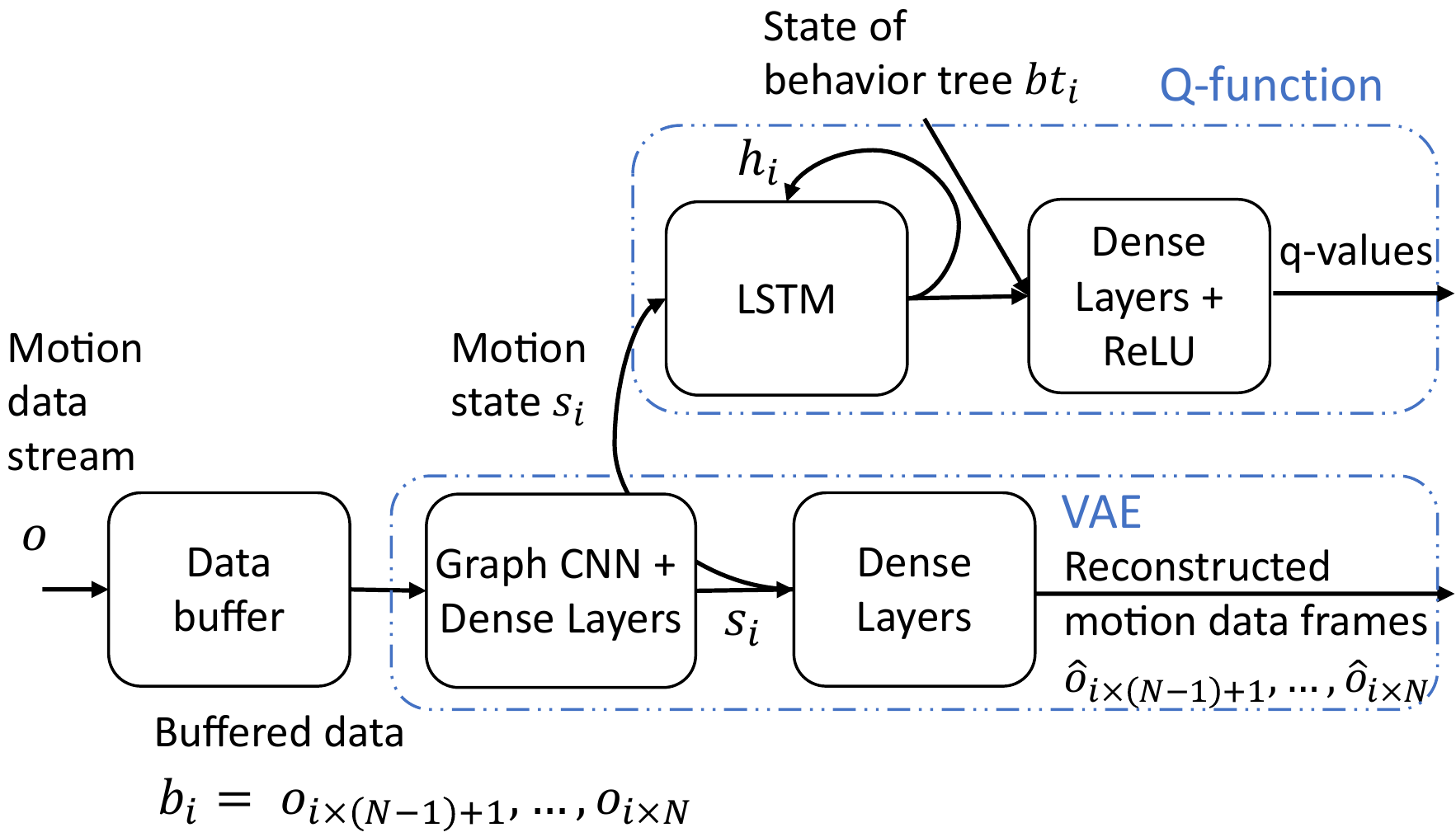}
\caption{The architecture of the learning framework. The architecture consists of a VAE structure to obtain a low-dimensional representation of the short-term motion data $s_i$ and a recurrent Q-function based on LSTM to output the state-action values.}
\label{fig:framework}
\end{figure}

The framework trains a model that extracts a representation of the short-term motion data and a recurrent deep Q-function.
As shown in Fig.~\ref{fig:framework}, the model consists of several layers of GCNs, long short-term memory (LSTM) blocks, and some dense, fully connected layers. GCNs process a fixed number of most recent motion data frames (25 frames) by several convolutional layers (4 layers) and a few dense layers (1 layer) and yields a compact low-dimensional short-term representation of the  motion data. This representation is fed into an LSTM block that outputs a more long-term representation of the motion data. The representation given by the LSTM layers is then processed by some dense layers (2 layers) to output the state-action value outputs. 

\subsection{Representation learning}
\label{sec:representation_learning}
Obtaining a compact representation of the input data is a common approach in policy training \cite{ghadirzadeh2017deep, chen2019adversarial, hamalainen2019affordance}.
Similarly, to improve the efficiency of the learning task, we learn a compact representation of several consecutive frames of motion data by training an auxiliary motion reconstruction task. 
The reconstruction network consists of an encoder $q(s_i | b_i)$ that assigns distributions over a low-dimensional latent variable $s_i$ conditioned on $N$ consecutive motion frames in the buffer $b_i = [o_{N \times (i-1) + 1}, ..., o_{N \times i}]$, and a decoder that reconstruct the original motion data given the latent variable as the input. 

As shown in Fig.~\ref{fig:framework}, the encoder consists of several layers of GCNs, followed by some dense layers. 
We exploit spatial-temporal graph convolutional networks (ST-GCN) \cite{yan2018spatial} in which the motion data is converted to an undirected graph based on the connectivity of the human body structure and the temporal relation of the frames. This representation of motion data is processed in a hierarchical structure by several convolutional layers followed by dense layers to form the low-dimensional latent representation $s_i$. 
The encoder and the decoder are trained end-to-end using
beta variational autoencoders ($\beta$-VAEs) \cite{higgins2017beta} to obtain a proper low-dimensional representation of the input data. 

VAEs approximate the likelihood function $p(b_i|s_i)$ and the posterior distribution $q(s_i|b_i)$ by optimizing the variational lower bound:
\begin{equation}
\mathcal{L}_{vae} = \mathbb{E}_{q(s_i|b_i)}[\log p(b_i|s_i)] 
 -\beta D_{KL}(q(s_i|b_i) || p(s_i)),
\label{eq:vae}    
\end{equation}
where $D_{KL}$ denotes the KL-divergence, $p(s_i)$ is the prior distribution over the latent variable, typically a normal distribution, and the parameter $\beta$ is a variable that controls the trade-off between the reconstruction fidelity and the distance between the posterior and the prior distribution \cite{higgins2017beta}. 
The VAE model is trained prior to the Q-learning task, and the representation that is found by the encoder model of the VAE is kept fixed and used to provide the inputs to the Q-function during the pre-training phase of the Q-function. 

\subsection{Deep recurrent Q-learning}
\label{sec:method:q-learning}
In this section, we describe the construction of the state-action value function based on the deep Q-network (DQN) method \cite{mnih2015human}, the way it is integrated in the behavior tree, and the generation of simulated data to train the function. 

We approximate the state-action value function by a deep recurrent  neural network $Q(\sigma_i, bt_i, a_i ; \theta_q)$, 
where $\theta_q$ denotes the parameters of the Q-function network, and $\sigma_i =  [s_0, ... s_i]$ denotes a full history of the states of the human motion until step $i$ and is processed sequentially by the LSTM layer of the Q-function.

At each training iteration, an experience $e_i = (\sigma_{i+1}, a_i, r_i,  bt_i)$ is sampled uniformly from a replay buffer $\mathcal{D}$.  The experience $e_i$ is used to optimize the following objective function: 
\begin{equation}
    \mathcal{L} = \mathbb{E}_{e_i \sim \mathcal{D}} [(r_i + \gamma \max_{a'} Q(\sigma_{i+1}, bt_i, a' ; \theta_q^-)
     - Q(\sigma_i, bt_i, a_i ; \theta_q))^2 ],
\end{equation}
where, $\theta_q^-$ denotes the previous parameter set of the network, and $\gamma$ is the discount factor. 
\subsubsection{Integrating the Q-function into the behavior tree}
As illustrated in Fig.~\ref{fig:bt1}B, the Q-function is integrated as a new node in the behavior tree to make proactive action-decisions. 
The Q-function receives the most recent motion data frames, as well as the state of the behavior tree (Fig.~\ref{fig:framework}), and recurrently updates its internal state and outputs the state-action values. 
For every action from the set, a trained Q-function must assign a value that is proportional to the expected amount of waiting time that will be eliminated by taking the action at the given time-step compared to the case when the robot does not take any proactive action.

The state of a node of a behavior tree can be either \textit{running}, \textit{success} or \textit{failure} \cite{iovino2020survey}. 
The Q-function node returns \textit{failure} when it needs more data to make a decision, i.e., when the action \textit{"wait to collect more data"} has the highest value as the output of the Q-function. 
The Q-function node is activated when an action other than \textit{"wait to collect more data"} has a high positive value. The node changes its status to \textit{running} while the action is being performed and upon a successful completion of the action, the node returns \textit{success}. 

\subsubsection{Simulation of high-level human-robot interaction}
The Q-function is updated in a trial-and-error manner based on the epsilon-greedy approach. 
It requires a large amount of data that may not be easy to collect in real human-robot collaborative setups. 
Therefore, we propose a simulation engine that simulates the effect of high-level robot action-decisions provided real motion data collected from several sessions of human-human collaborations. 
It is important to note that the simulator only simulates the high-level interactions between the human and the robot, and the policy that is trained in such an environment can be directly transferred to the real task setup.
\begin{figure*}[h]
    \centering
    \includegraphics[width=1\textwidth]{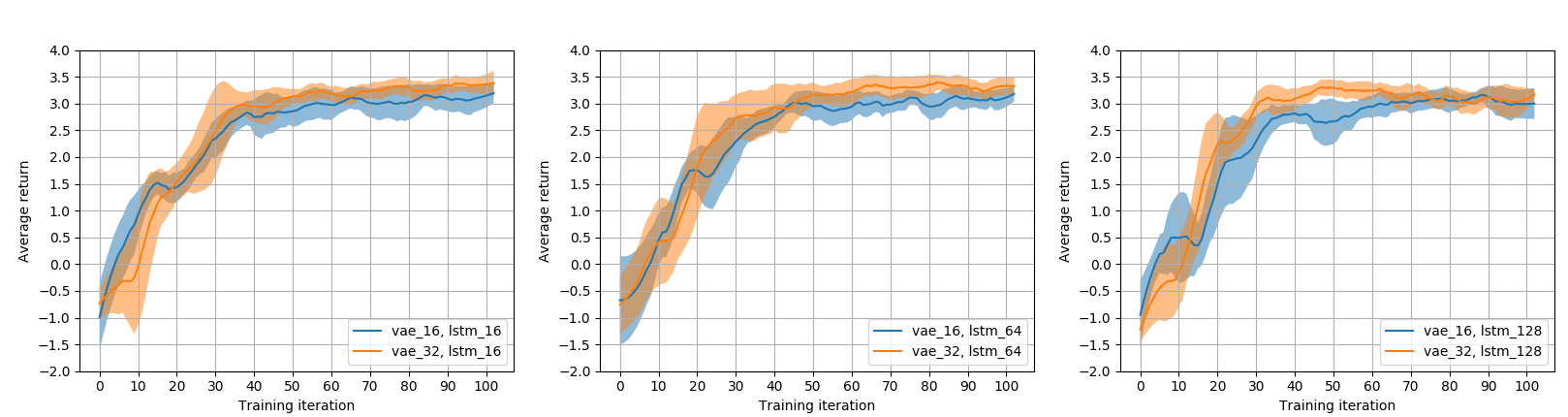}
    \caption{The learning curve with different VAE latent dimensions and LSTM hidden sizes. The models are trained independently for 5 times. The mean and standard deviation of the learning curves are presented in the figure.}
    \label{fig:learning_curve}
\end{figure*}

\section{Experiments}
\label{sec:experiments}

In this section, we present experimental results aimed at answering the following questions:
\begin{enumerate}
    \item Does the proposed learning framework permit more fluent coordination between human and robot partners in collaborative setups?  
    \item Does the exploitation of motion sensing devices, in particular wearable motion capture suits, enhance the perception of a robot to better recognize human activities and behave more proactively?
    \item Does the proposed architecture capture a proper representation of long-term human behavior suitable for robot policy training? How do the parameters of the architecture affect the performance? 
    \item How well does end-to-end RL training of the perception system and policy perform compared to traditional methods of recognizing human activities in isolation from robot policy training? 
\end{enumerate}

We answer the first three questions by using the proposed learning framework in a collaborative packaging task, and demonstrate that the efficiency of the coordination between the human-robot pair improves when the robot takes proactive decisions. We also train the model with different parameters of the proposed architecture to properly answer (3). 
Finally, in order to answer (4) we bench-mark our method against a more traditional approach to implement proactive behavior \cite{zhang2020recurrent,kwon2014planning,hawkins2014anticipating,huang2016anticipatory}.

\begin{table}[h]
    \centering
    \caption{Reward values estimated based on the performance of the YuMi robot to complete the given packaging task.}
    \begin{tabular}{l|l|l}
         \textbf{Human action }      & \textbf{Robot action}  & \textbf{Reward}\\
         \hline
        picks up, carries     & pick up right item    & $\min(t_{b}^*, 4.0)$\\
        and places a     & and right position  & \\
        \cline{2-3}
        box on the table & pick up right item   & $\min(t_{b}, 3.5)$\\
                           & but wrong position&       \\
        \cline{2-3}                   
                           & pick up wrong item   &   $\max(-t_{b}, -3.5)$ \\ 
        \hline
        picks up and & pick up right item & $\min(t_{b}, 3.5)$\\
        \cline{2-3}
        places air & pick up wrong item & $\max(-t_{b}, -3.5)$\\
        \cline{2-3}
        bubble sheets & lift the box & $\max(-t_{b}, -4.5)$\\
        \hline
        wrapping up & lift the box & $\min(t_{b}, 4.5)$ \\
        \cline{2-3}
         &pick up an item & $\max(-t_{b}, -3.5)$ \\
         \hline
    \end{tabular}
    \vspace{1ex}
    
     {\raggedright \quad\quad\quad\quad\quad\quad\quad\quad\quad\quad\quad\quad$^* t_b$ is the time to get information from the barcode reader \par}
    \label{tabel:exp:rewards}
\end{table}

\subsection{Task setup}
\label{sec:exp:setup}
We evaluate the learning framework on a collaborative packaging task, shown in Fig.~\ref{fig:setup}, in which a person wears a Rokoko motion capture suit and collaborates with an ABB YuMi robot. In this task scenario, the person chooses between two types of boxes, picks up a box, and moves it to the robot. The box is arbitrarily placed on either of the two alternative positions in front of the robot. Once the box is on the table, its barcode is scanned and the position and the type of the box are sent to the robot. Depending on the type, the robot has to pick up an item from the correct bin and put it inside the box. The person can then choose to ask for more items or wrap up the box. In the former case, the person puts an air bubble wrap sheet, and then the robot puts another item into the box. In the latter case, the person picks up the wrap tape, and the robot lifts the box to help the person to complete the wrapping. 

As described earlier, the reward is given according to the amount of time that is saved by making proactive action decisions. Based on our experimental results and the time that is required by the robot to perform a task, e.g., to pick and place an item, estimated rewards in different conditions are given in TABLE \ref{tabel:exp:rewards}.

\subsection{Learning performance}
\label{sec:exp:learning_performance}
In this section, we provide the training performance for the proposed learning framework. 
We collected 200 sessions of human-human collaboration data for the joint packaging task. 
The data is used to train the VAE model, and also to update the state-action value function using a simulator that uses the rewards given in Table \ref{tabel:exp:rewards}.
The value function is updated for 100 iterations using the recurrent DQN method. Each training iteration consists of 50 mini-batch updates, and each batch contains 128 experiences sampled uniformly from the replay buffer. 

The learning performance of different network architectures is presented in Fig.~\ref{fig:learning_curve}. The figure illustrates the average of five independent training trials for every combination of network configurations. 
We evaluated the learning framework, and the effect of the important parameters of the network, i.e., the size of the latent space of the  VAE, and the number of hidden neurons of the LSTM. 

As shown in Fig.~\ref{fig:learning_curve}, the learning framework successfully improves the time-efficiency of the task. 
The trained models enable the robot to make proactive action-decisions which on average eliminate 3.4 seconds of the waiting time in each phase of the collaboration. 
In our task setup, a typical packaging session consists of six such phases. 
Besides, we observed that the number of hidden neurons of LSTM has little impact on the learning performance. However, the size of the VAE latent space slightly affects the learning performance. In general, we conclude that the performance of the learning framework is not considerably influenced by the choice of the number of parameters.

In order to evaluate the quality of the features that are extracted by the proposed VAE method, we trained a different state representation by introducing a supervised auxiliary loss in addition to the reconstruction loss. 
We annotated the human motion data by labeling different activities, and trained an auxiliary classifier that consists of the encoder model and some auxiliary dense layers. The classifier is trained jointly with the VAE model to extract a new representation that contains information to reconstruct the motion data and to classify human activities. 

\begin{figure}[h]
    \centering
    \includegraphics[width=0.6\textwidth]{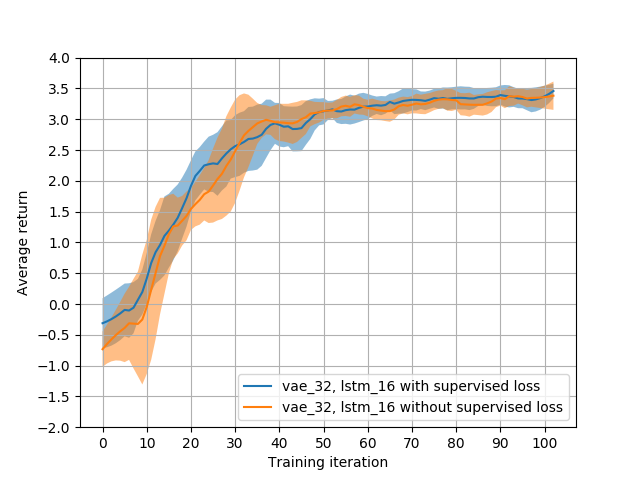}
    \caption{Comparison of the learning performance with and without the supervised loss to construct the low-dimensional representation of the motion data. The models are trained independently for 5 times. The mean and standard deviation of the learning curves are presented in the figure.}
    \label{fig:supervised_rep}
\end{figure}

Fig.~\ref{fig:supervised_rep} illustrates the Q-learning performance based on the representation constructed with the additional supervised auxiliary task, and compares similarly to the performance to the original architecture.
We conclude that the original representation constructed based on only the unsupervised auxiliary task is rich enough for the Q-learning task, and adding extra supervision does not improve the performance. 

\begin{figure}[h]
    \centering
    \includegraphics[width=0.6\textwidth]{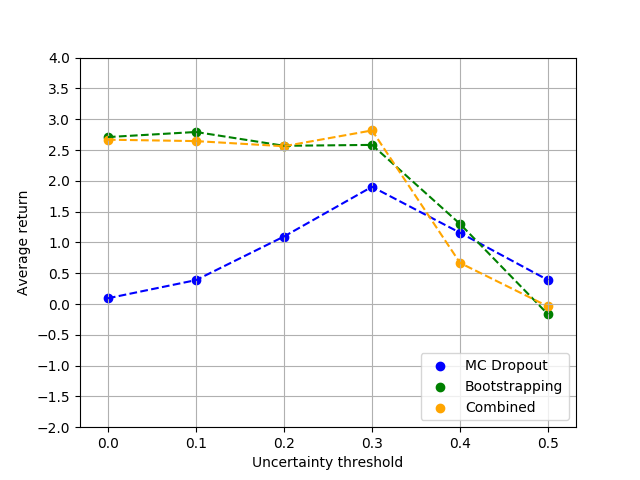}
    \caption{The average reward of the three baseline methods for different threshold values.}
    \label{fig:bench_mark}
\end{figure}

\subsection{Bench-marking}
\label{sec:exp:benchmark}
We compare our proposed RL framework to a baseline method that trains a supervised classifier. Given the motion data as input, the classifier outputs the right robot action and a measure of uncertainty at every time-step. When the uncertainty is higher than a threshold, the output of the model is discarded and the command \textit{"wait to collect more data"} is executed. 
Otherwise, the action given by the output of the model is executed. 

The new architecture is similar to the architecture shown in Fig.~\ref{fig:framework} with two differences: (1) the output of the network is given by a arg-softmax operator, and (2) a dropout unit is added to the output of the final dense layer. 
The entire network is trained end-to-end with supervised learning by providing the correct action labels for every piece of motion data. The new node is integrated in the behavior tree similar to the state-action value function. 

Following the work in \cite{kahn2017uncertainty}, we obtain a measure of uncertainty based on (1) bootstrapping, (2) dropout, and (3) the combination of dropout and bootstrapping. We evaluate the efficiency of the given task on a range of threshold values to obtain the best possible performance. 
As it is illustrated in Fig.~\ref{fig:bench_mark}, 
our proposed RL method outperforms  the baseline approaches in terms of the average reward. 
The best performance is achieved by combining the bootstrapping and dropout techniques and setting the threshold to 0.3. In this case, the average reward is about 2.9 seconds which is considerably lower than the average reward given by the proposed RL framework (3.4 seconds). 
Besides, please note that the proposed RL framework does not require any supervision, while the baseline methods are trained using annotated motion data. 

\section{CONCLUSIONS}
\label{sec:conclusion}

For a collaborative robot to be effective when engaged in joint tasks with a human, it has to proactively contribute to the task without the knowledge of the full state of the system, including the human and the environment. The robot should be able to act as soon as it has gathered enough confidence that its own action will contribute to execute the task faster.

We have proposed a RL based framework that effectively deals with the uncertainties in perception and finds an optimal balance between timely actions and the risk of making mistakes. Our experiments show that this permits for a more fluent coordination between human and robot partners since unnecessary delays can be avoided. We have also shown that, compared to the unsupervised learning paradigm used by the proposed framework, the benefit of an additional supervised learning loss is limited. In practice, this means that tedious annotation of motion data can be avoided, which makes it easier for the framework to be retrained to novel tasks. Future research will thus be focused on faster adaptation to new human partners, partners that all behave somewhat differently even for the same task, and faster transition from one execution plan to other potentially more complex ones. 

\section*{Acknowledgments}
This work was supported by Knut and Alice Wallenberg Foundation, the EU through the project EnTimeMent and the Swedish Foundation for Strategic Research through the COIN project.

\bibliographystyle{IEEEtran}
\bibliography{references}  
\end{document}